# Design of a Spherical Wrist with Parallel Architecture: Application to Vertebrae of an Eel Robot

Damien Chablat and Philippe Wenger

*Institut de Recherche en Communications et Cybernétique de Nantes*
*1, rue de la Noë, BP 92101, 44321 Nantes Cedex 03 France*
Damien.Chablat@irccyn.ec-nantes.fr

*Abstract* - **The design of a spherical wrist with parallel architecture is the object of this article. This study is part of a larger project, which aims to design and to build an eel robot for inspection of immersed piping. The kinematic analysis of the mechanism is presented first to characterize the singular configurations as well as the isotropic configurations. We add the design constraints related to the application, such as (i) the compactness of the mechanism, (ii) the symmetry of the elements in order to ensure static and dynamic balance and (iii) the possibility of the mechanism to fill the elliptic form of the ell sections.**

## I. INTRODUCTION

Over millions of years, fish have evolved swimming capacity far superior in many ways to what has been by nautical science and technology. They use their streamlined bodies to exploit fluid-mechanical principles. This way, they can achieve extraordinary propulsion efficiencies, acceleration and maneuverability not feasible by the best naval architects [1].

Parallel kinematic architectures are commonly claimed to offer several advantages over their serial counterparts, like high structural rigidity, high dynamic capacities and high accuracy [2]. Thus, they are interesting for applications where these properties are needed, such as flight simulators [3] and high-speed machines. Recently, new applications have used such mechanisms to build humanoid robots [4], or snake robots [5].

The purpose of this article is to design vertebrae of an eel robot by using the advantages of the parallel architectures while approaching eel morphology. The next section presents the objectives of the biomimetic as well as suitable spherical architectures. The design parameters and the kinematics of the mechanism to be optimized are reported in Section 3.

## II. PRELIMINARIES

### A. Biomimetic robotics

The object of the biomimetic robotics is to mimic life, to imitate biological systems or to conceive new technologies drawn from the lesson of their study [6].

For the last twenty years or so, many researches have been made in the underwater field in America and in Japan [7]. Among those, a good number attempted to reproduce fish. In this context, two modes of locomotion mainly attract the attention of researchers (i) the carangid swimming (family Carangidae as jacks, horse mackerel or pompano [8]) based on oscillations of the body and (ii) the anguilliform swimming (of snake type, eel, lamprey, etc.) based on undulations of the body. An anguilliform swimmer propels itself forward by propagating waves of curvature backward along its body [1].

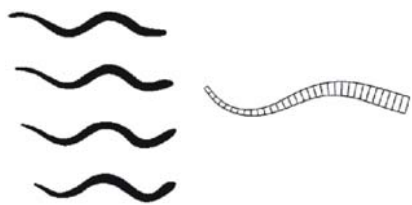

Fig. 1: Change in body shape in swimming and a subdivision of its body

To carry out anguilliform swimming, the body of the eel is made of a succession of vertebrae whose undulation produces motion, as depicted in Fig. 1. In nature, there is only one degree of freedom between each vertebra because the motion control of the vertebrae is coupled with the motion of the dorsal and ventral fin. These two fins being not easily reproducible, we will give to each vertebra, more mobility to account problems of rolling, for example. The assembly of these vertebrae, coupled to a head having two fins must allow the reproduction of the eel swimming.

From the observation of European eel, Anguilla anguilla, we have data concerning his kinematic swimming such as wave speed, cycle frequency, amplitude or local bending [9]. The yaw is given for forward and backward swimming on total body length, as depicted in Fig. 2. The other angles are obtained using Navier-Stokes equations on characteristic trajectories [10]. For our prototype, we took as constraints of design, ±30 degrees in yaw for forward swimming, ±15 degrees in pitching for diving and ±4 degrees in rolling to compensate for torsion in diving.

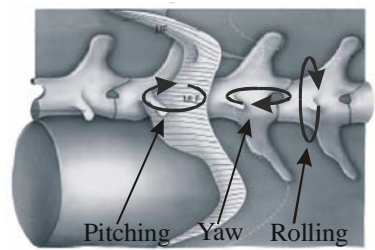

Fig. 2: Rolling, pitching and yaw angles of vertebrae

The objective of our study is to build an eel robot with 10 vertebrae and an overall length of 1500 mm (with the head and the tail included). Each vertebra will have an elliptic section of 150 mm and 100 mm focal distance respectively and will be a 100 mm thick.

### B. Mechanical architectures

The design of the vertebrae of an eel is equivalent to the design of spherical wrists. Indeed, an eel being able to be comparable with a beam, two theories can govern its moves, (i) the theory of Reissner [11], which imposes a 6 DOF kinematic mechanism (realizable by the stacking of Gough-Steewart's platforms) and (ii) the theory of Kirchoff of the inextensible beams, which imposes the kinematics of the ball joint type [12].

Spherical wrists with serial architecture raise several problems, which lead us to study parallel architectures. The first problem is related to the compactness of the wrist because the distance between the successive vertebrae must be minimized in order to create a continuous deformation of the eel body. The second problem is related to the singularity of serial wrist (first and last axis aligned) and, if we use such architecture, only the second revolute joint is mainly used to produce the oscillation of the body, which yields problems of dimension.

Conversely, spherical wrists with parallel architecture are numerous [3-13]. If we want to mimic the displacements created by the muscles, the corresponding joint is a prismatic actuator. Most existing actuators are based on the use of a rotary motor, a reducer and a ball screw. Such devices are used, for example, in the flight simulators using Gough-Stewart's platforms. In the next section, we will present an architecture that uses revolution joints to produce equivalent motions.

### III. KINEMATIC STUDY OF THE SELECTED ARCHITECTURE

#### A. Description

Spherical parallel mechanisms can be classified into two main groups, symmetrical or asymmetrical mechanisms, which can be overconstraint or non-overconstraint [14].

From the first group, we have the agile eye [15] which uses revolute joints fixed on the base (Fig. 3). It is firstly developed for the rapid orientation of a camera but it is also used for carrying a tool [16].

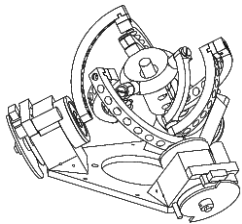

Fig. 3: The agile eye [15]

For example, the miniature camera attached to the end-effector can be pointed in a cone of vision of 140° with ±30° in torsion [17]. Such properties are not asked for our prototype because only the yaw angle must be higher and it is difficult to place the actuated joints on an elliptic basis. Thus, we will study a spherical wrist, which can produce high yaw and where engine torques can be added by using the principle of the differential mechanism.

The selected architecture is a non-overconstrained asymmetrical architecture that is reported in [14] as an (3, 6, 6) architecture. The base and the mobile platform are connected by three kinematic chains, as depicted in Fig. 4.

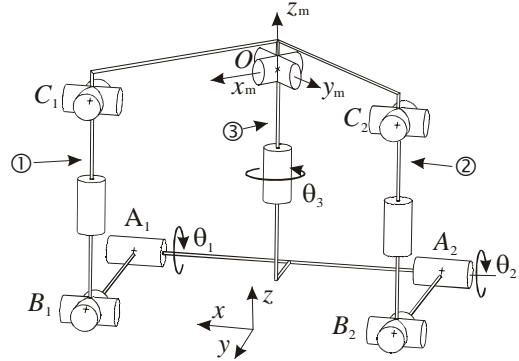

Fig. 4: Structure of the studied spherical wrist

This architecture results from the research around the Lie Group of Euclidian displacements [13]. There are (i) two kinematic chains, noted legs ① and ②, to produce a general rigid body displacement from the subgroup {D} (6 DOF) and (ii) a kinematic chain, noted leg ③, from the spherical subgroup {S} and made by three coaxial revolute joints (3 DOF). There is only one actuated joint on each leg $(\theta_1, \theta_2, \theta_3)$.

If the realization of leg ③ is easy (three coaxial revolute joint), it is difficult to enumerate all the legs with 6-DOF. The most current generator of {D} is of the UPS type (Gough-Stewart's platform, with P prismatic actuated joint, U for universal joint and S for spherical joint), which has the disadvantage of using a prismatic actuator that is not fixed on the basis.

In the literature, an equivalent mechanism exists but the generator of {D} is of PUS type (with P prismatic actuated joint). For legs ① and ②, the prismatic actuated joints are in parallel to the vertebral column which is harmful for the compactness of the mechanism. The orientation can be changed but the efficiency decreases considerably. For leg ③, the first revolute joint (located on the base) is actuated.

Thus, we have changed the type of legs ① and ②, by a RUS type (with R revolute actuated joint) as depicted in Fig. 4. On the next section, we will justify their placements and their dimensions by the study of the Jacobian matrix.

#### B. Kinematics

A fixed reference frame, noted $\Re_{fixed}(O, x, y, z)$ is located on the base and is oriented in such a way that (i) plane $Oxz$ is planed by points $C_1$, $C_2$ and $O$, (ii) the z-axis is vertical, (iii) x-axis is directed from $A_2$ to $A_1$. The coordinates of points $A_1$ and $A_2$ in $\Re_{fixed}$ are written as

$$[OA_1]_{\Re_{fixed}} = \begin{bmatrix} a & b & c \end{bmatrix}^T \text{ and } [OA_2]_{\Re_{fixed}} = \begin{bmatrix} -a & b & c \end{bmatrix}^T \quad (1)$$

The lengths $a$, $b$, and $c$ will be chosen by the study of the Jacobian matrix in the next subsection.

The mobile platform will be rotating around point $O$ that is the origin of the mobile frame, noted $\Re_{mobile}$. The orientation of $\Re_{mobile}(O, x_m, y_m, z_m)$ is defined so that (i) plane $Ox_m y_m$ is the plane defined by points $C_1$, $C_2$ and $O$, (ii) $x_m$-axis is directed from $O$ to $C_1$ and (iii) $y_m$-axis is directed from $O$ to $C_2$.

Let $\theta$ be the vector of joint coordinates associated with the actuated revolute joints. The orientation of the mobile platform with respect to fixed frame $\Re_{base}$ is defined by the "Rolling Pitching Yaw" parameters (RPY) where the first parameter is the orientation angle $\theta_3$ of the first revolute joint of leg ③).

$$\theta = \begin{bmatrix} \theta_1 & \theta_2 & \theta_3 \end{bmatrix}^T$$

$$^{fixed}\mathbf{R}_{mobile} = \mathbf{R}(z, \theta_3)\mathbf{R}(y', \phi)\mathbf{R}(x'', \psi)$$

The angles $\theta_3$, $\phi$ and $\psi$ are associated with the following cascaded rotations

(i) a rotation of angle $\theta_3$ around $z$-axis,

(ii) a rotation of angle $\phi$ around the $y'$-axis (obtained from the previous rotation and whose axis is the axis of the second revolute joint of leg ③),

(iii) a rotation of $\psi$ around the $x''$-axis (obtained from the second rotation and whose axis is the axis of the third revolute joint of leg ③).

### C. Jacobian matrices

To characterize the singular configurations, we will use an invariant form, which allows our results to be applicable to any architecture studied here. Thus, there is no problem of singularity of transformation in the rotation matrix between $\Re_{fixed}$ and $\Re_{mobile}$.

We write the Chasles's relation on $(\mathbf{c}_i - \mathbf{b}_i)$ to have

$$(\mathbf{c}_i - \mathbf{b}_i) = (\mathbf{c}_i - \mathbf{o}_i) + (\mathbf{o}_i - \mathbf{a}_i) - (\mathbf{b}_i - \mathbf{a}_i) \quad (2)$$

In this equation, all the vectors are expressed in $\Re_{fixed}$. To simplify calculations, we set

$$\mathbf{r}_i = (\mathbf{c}_i - \mathbf{b}_i), \ \mathbf{p}_i = (\mathbf{c}_i - \mathbf{o}_i), \ \mathbf{b}_i = (\mathbf{o}_i - \mathbf{a}_i) \text{ et } \mathbf{l}_i = (\mathbf{b}_i - \mathbf{a}_i)$$

By differentiating Eq. (2) with respect to time, we obtain,

$$\dot{\mathbf{r}}_i = \dot{\mathbf{p}}_i - \dot{\mathbf{l}}_i \quad (3)$$

with

$$[\mathbf{p}_i]_{\Re_{fixed}} = {}^{fixed}\mathbf{R}_{mobile}[\mathbf{p}_i]_{\Re_{mobile}} \quad (4)$$

Differentiating with respect to time, we find

$$[\dot{\mathbf{p}}_i]_{\Re_{fixed}} = \dot{\mathbf{Q}}\,{}^{fixed}\mathbf{R}_{mobile}[\mathbf{p}_i]_{\Re_{mobile}} \quad (5)$$

since vector $[\mathbf{p}_i]_{\Re_{mobile}}$ is a constant vector when expressed in frame $\Re_{mobile}$. Moreover, the time derivation of the rotation matrix can be written as

$$\dot{\mathbf{Q}} = \mathbf{\Omega}\mathbf{Q} \quad (6)$$

where $\mathbf{\Omega}$ is the angular velocity tensor. Finally, from Eqs. (2) and (6), we get

$$\dot{\mathbf{p}}_i = \mathbf{\Omega}\mathbf{p}_i = \mathbf{\omega} \times \mathbf{p}_i$$

where $\times$ denotes the cross product of the two vectors and $\mathbf{\omega}$ is the angular velocity vector. We note $\mathbf{i}_1$ and $\mathbf{i}_2$, the unit vectors passing through the axis of the first revolute joint of legs ① and ②, respectively. Moreover, we can write vector $\dot{\mathbf{l}}_i$ as function of angular velocities $\dot{\theta}_1$ and $\dot{\theta}_2$

$$\dot{\mathbf{l}}_1 = \mathbf{l}_1 \times (\dot{\theta}_1 \mathbf{i}_1) \text{ and } \dot{\mathbf{l}}_2 = \mathbf{l}_2 \times (\dot{\theta}_2 \mathbf{i}_2)$$

Thus, Eq. (3) can be written in the form

$$\dot{\mathbf{r}}_i = \mathbf{\omega} \times \mathbf{p}_i - \mathbf{l}_i \times (\dot{\theta}_i \mathbf{i}_i)$$

We multiply the preceding equation by $\mathbf{r}_i^T$ because $\mathbf{r}_i^T \dot{\mathbf{r}}_i = 0$. Thus, we have

$$\mathbf{r}_i^T.(\mathbf{\omega} \times \mathbf{p}_i) = \mathbf{r}_i^T.(\mathbf{l}_i \times (\dot{\theta}_i \mathbf{i}_i))$$

Or

$$(\mathbf{p}_i \times \mathbf{r}_i)^T.\mathbf{\omega} = (\mathbf{l}_i \times \mathbf{r}_i)^T.(\dot{\theta}_i \mathbf{i}_i)$$

These two equations can be cast in vector form

$$\mathbf{A}\mathbf{\omega} + \mathbf{B}\dot{\mathbf{q}} = 0 \quad (7)$$

with

$$\mathbf{A} = \begin{bmatrix} (\mathbf{p}_1 \times \mathbf{r}_1)^T \\ (\mathbf{p}_2 \times \mathbf{r}_2)^T \\ 0 & 0 & 1 \end{bmatrix} \quad (8)$$

$$\mathbf{B} = \begin{bmatrix} (\mathbf{l}_1 \times \mathbf{r}_1)^T.\mathbf{i}_1 & 0 & 0 \\ 0 & (\mathbf{l}_2 \times \mathbf{r}_2)^T.\mathbf{i}_2 & 0 \\ 0 & 0 & 1 \end{bmatrix} \quad (9)$$

and $\dot{\mathbf{q}} = \begin{bmatrix} \dot{\theta}_1 & \dot{\theta}_2 & \dot{\theta}_3 \end{bmatrix}^T$

Then, when $\mathbf{B}$ is not singular, the inverse Jacobian matrix is written,

$$\mathbf{J}^{-1} = \begin{bmatrix} \dfrac{1}{(\mathbf{l}_1 \times \mathbf{r}_1)^T.\mathbf{i}_1}(\mathbf{p}_1 \times \mathbf{r}_1)^T \\ \dfrac{1}{(\mathbf{l}_2 \times \mathbf{r}_2)^T.\mathbf{i}_2}(\mathbf{p}_2 \times \mathbf{r}_2)^T \\ 0 \quad 0 \quad 1 \end{bmatrix}$$

### D. Singular configurations

The parallel singularities occur when the determinant of the matrix $\mathbf{A}$ vanishes, *i.e.* when $\det(\mathbf{A}) = 0$. In such configurations, it is possible to move locally the mobile platform whereas the actuated joints are locked. These singularities are particularly undesirable because the structure cannot resist any force or torque.

Form Eq. (7), we have

$$(\mathbf{p}_1 \times \mathbf{r}_1) \square (\mathbf{p}_2 \times \mathbf{r}_2) \text{ or } (\mathbf{p}_1 \times \mathbf{r}_1) = 0 \text{ or } (\mathbf{p}_2 \times \mathbf{r}_2) = 0$$

It is equivalent to have $B_1$, $B_2$, $C_1$, $C_2$ and $O$ coplanar or to have ($B_1$, $C_1$, $O$) or ($B_2$, $C_2$, $O$) aligned, as depicted in Fig. 5.

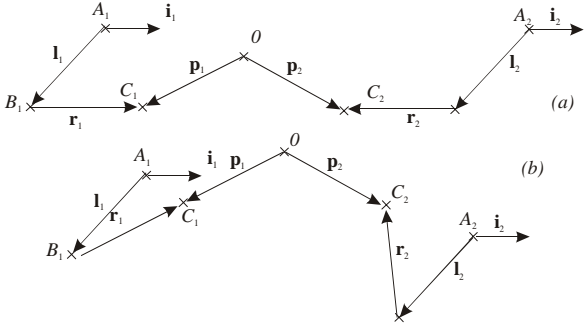

Fig. 5: Parallel singularity when
(a) $B_1$, $B_2$, $C_1$, $C_2$ and $O$ are coplanar and (b) $B_1$, $C_1$ and $O$ are aligned

Serial singularities occur when the determinant of the matrix **B** vanishes, *i.e.* when $\det(\mathbf{B}) = 0$. At a serial singularity, an orientation exists along which any angular velocity cannot be produced.

From Eq. (8), we have

$$(\mathbf{l}_1 \times \mathbf{r}_1)^T \mathbf{i}_1 = 0 \text{ or } (\mathbf{l}_2 \times \mathbf{r}_2)^T \mathbf{i}_2 = 0 \text{ or}$$

$$(\mathbf{i}_1 \times \mathbf{r}_1)^T \mathbf{l}_1 = 0 \text{ or } (\mathbf{i}_2 \times \mathbf{r}_2)^T \mathbf{l}_2 = 0$$

It is equivalent to have (i) $\mathbf{l}_1$ and $\mathbf{r}_1$ aligned, or (ii) $\mathbf{l}_2$ and $\mathbf{r}_2$ aligned, or (iii) $\mathbf{r}_1$ and $\mathbf{i}_1$ aligned, or (iv) $\mathbf{r}_2$ and $\mathbf{i}_2$ aligned, as depicted in Fig.6.

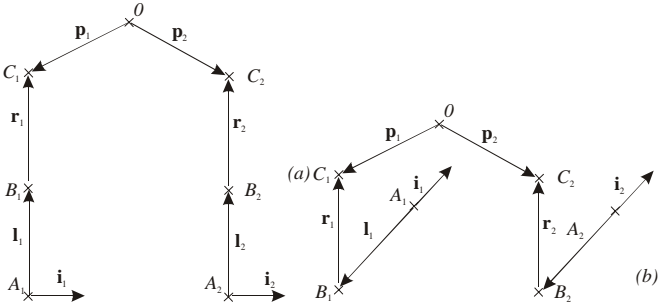

Fig. 6: Serial singularity when
(a) $\mathbf{l}_i$ and $\mathbf{r}_i$ are aligned and (b) $\mathbf{l}_i$ and $\mathbf{i}_i$ are aligned

### E. Condition number and isotropic configurations

The Jacobian matrix is said to be isotropic when its condition number attains its minimum value of one [18]. The condition number of the Jacobian matrix is an interesting performance index, which characterises the distortion of a unit ball under the transformation represented by the Jacobian matrix.

The Jacobian matrix of a manipulator is used to relate (i) the joint rates and the Cartesian velocities, (ii) the static load on the output link and the joint torques or forces. Thus, the condition number of the Jacobian matrix can be used to measure the uniformity of the distribution of the tool velocities and forces in the Cartesian workspace.

The aim of this section is to define the placement and dimensions of each vertebra to eliminate the singular configurations from the workspace and to maximize the kinematic properties around its isotropic configurations.

The matrix **A** is isotropic when

$$\mathbf{p}_1 \perp \mathbf{r}_1 \text{ and } \mathbf{p}_2 \perp \mathbf{r}_2 \text{ and}$$

$$(\mathbf{p}_1 \times \mathbf{r}_1) \perp (\mathbf{p}_2 \times \mathbf{r}_2) \text{ and } \|\mathbf{r}_1\| = \|\mathbf{p}_1\| = \|\mathbf{r}_2\| = \|\mathbf{p}_2\| = 1$$

Moreover, **A** is equal to the identity matrix, as depicted in Fig. (7). The cross product $(\mathbf{p}_1 \times \mathbf{r}_1)$ and $(\mathbf{p}_2 \times \mathbf{r}_2)$ and the *z*-axis form an orthogonal frame.

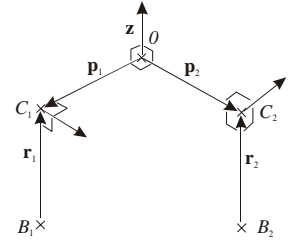

Fig. 7: Isotropic configuration of matrix A

The matrix **B** is isotropic and equal to the identity matrix when

$$\mathbf{l}_1 \perp \mathbf{r}_1 \text{ and } \mathbf{l}_2 \perp \mathbf{r}_2 \text{ and } \mathbf{l}_1 \perp \mathbf{i}_1 \text{ and } \mathbf{l}_2 \perp \mathbf{i}_2 \text{ and } \|\mathbf{r}_i\| = \|\mathbf{l}_i\| = 1$$

The result of the isotropic constraints on **A** and **B** are shown in Fig. 4 as an example. In fact, there is an infinity of solution because no constraint gives us the orientation of $\mathbf{i}_1$ compared to $\mathbf{i}_2$. Figure 4 shows $\mathbf{i}_1$ and $\mathbf{i}_2$ parallel but they can be different.

### F. Biomimetic constraints and model simplifications

From the preceding result, we will present three suitable solutions of our prototype.

• The first solution is the mechanism depicted in Fig. 4 that we could call "parallel axes". Equation 1 gives the location of points $A_1$ and $A_2$ in $\Re_{base}$ for a unit mechanism,

$$a = \frac{\sqrt{2}}{2}, \quad b = \frac{\sqrt{2}-2}{2}, \quad c = -1$$

If this solution admits an isotropic configuration, the behaviour in forward swimming leads to use legs ① and ② simultaneous. When we apply as input velocity $\dot{\boldsymbol{\theta}} = \begin{bmatrix} 1 & 1 & 0 \end{bmatrix}^T$, the angular velocity obtained is $\boldsymbol{\omega} = [\sqrt{2} \ 0 \ 0]^T$. This means that we amplify the rotational motion just after having used a reduction gear on the rotary motor to increase the available torque. Thus, the length of the motors is constrained by the shape of the cross section of the eel, as depicted in Fig. 9 (a).

• The second solution, called "orthogonal axes", is to place $\mathbf{i}_1$ and $\mathbf{i}_2$ orthogonally as depicted in Fig. 8. The location of points $A_1$ and $A_2$ in $\Re_{base}$ coincides with point $O$.

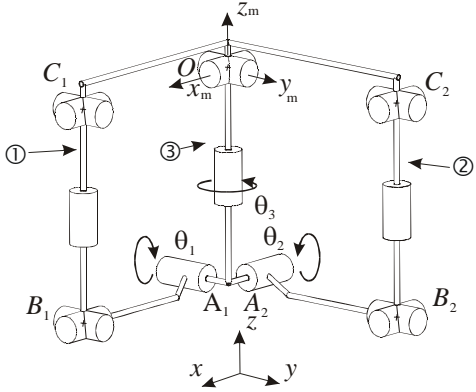

Fig. 8: Spherical wrist with orthogonal actuators

In this case, the direct and inverse kinematic models are simpler but it is more difficult to place the motors of legs ① and ②, as shown in Fig. 9 (b). Moreover, there also exists an angular amplification factor in the forward swimming.

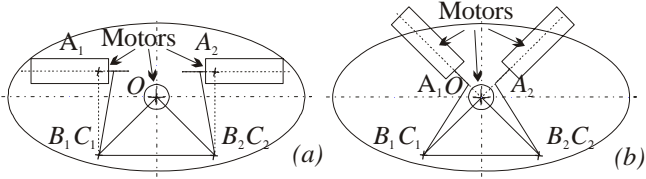

Fig. 9: Placement of the motors and the legs for
(a) the "parallel axes" and (b) the "orthogonal axes"

• The last solution has parallel actuators and their axes intersect the z-axis, as depicted in Fig. 10. When the eel robot is swimming, the angular velocity of the actuated joints of legs ① and ② is equal to yaw velocity.

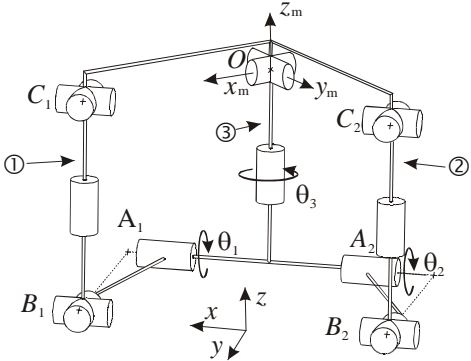

Fig. 10: Spherical wrist with parallel actuators

This means that for the forward or backward swimming, the kinematic models are simple and the torque needed for the motion is distributed. However, only **A** can be isotropic because we have

$$(\mathbf{l}_i \times \mathbf{r}_i)^T \cdot \mathbf{i}_i = \sqrt{2}/2 \quad \text{for i=1,2}$$

Equation 1 gives the location of points $A_1$ and $A_2$ in $\Re_{base}$ for a unit mechanism,

$$a = \sqrt{2}/2, \quad b = 0, \quad c = -1$$

Concerning the integration into the cross-section of the eel, the placement is less constrained, as shown in Fig. 11.

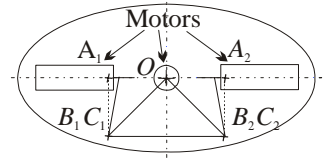

Fig. 11: Placement of the motors and the legs for spherical wrist with parallel actuators

### G. Direct and inverse kinematic models

The direct kinematic model can be written when we know the position of $B_i$ and $C_i$. Thus, we have,

$$B_1 = \left[\frac{\sqrt{2}}{2} \quad \frac{\sqrt{2}C_1}{2} \quad -1+\frac{\sqrt{2}S_1}{2}\right]^T \quad B_2 = \left[-\frac{\sqrt{2}}{2} \quad \frac{\sqrt{2}C_2}{2} \quad -1+\frac{\sqrt{2}S_2}{2}\right]^T$$

and in $\Re_{mobile}$,

$$C_1 = \begin{bmatrix} 1 & 0 & 0 \end{bmatrix}^T \quad C_2 = \begin{bmatrix} 0 & 1 & 0 \end{bmatrix}^T$$

or in $\Re_{fixed}$

$$C_1 = \begin{bmatrix} C_3 C_\phi \\ S_3 C_\phi \\ -S_\phi \end{bmatrix} \quad C_2 = \begin{bmatrix} C_3 S_\phi S_\psi - S_3 C_\psi \\ S_3 S_\phi S_\psi + C_3 C_\psi \\ C_\phi S_\psi \end{bmatrix}$$

with $C_i = \cos(\theta_i)$, $S_i = \sin(\theta_i)$ for i=1,2,3, $C_\phi = \cos(\phi)$, $S_\phi = \sin(\phi)$, $C_\psi = \cos(\psi)$ and $S_\psi = \sin(\psi)$.

We add the constraint that $\|B_i C_i\| = 1$

$$\left|C_3 C_\phi - \frac{\sqrt{2}}{2}\right|^2 + \left|S_3 C_\phi - \frac{\sqrt{2}}{2}C_1\right|^2 + \left|S_\phi - 1 + \frac{\sqrt{2}}{2}S_1\right|^2 = 1 \quad (10)$$

$$\left|C_3 S_\phi S_\psi - S_3 C_\psi + \frac{\sqrt{2}}{2}\right|^2 + \left|S_3 S_\phi S_\psi + C_3 C_\psi - \frac{\sqrt{2}}{2}C_2\right|^2 + \left|\frac{\sqrt{2}}{2}S_2 - C_\phi S_\psi - 1\right|^2 = 1 \quad (11)$$

To solve the direct kinematic, we know $\boldsymbol{\theta} = \begin{bmatrix} \theta_1 & \theta_2 & \theta_3 \end{bmatrix}^T$ and we use the following substitutions

$$\sin(\phi) = \frac{2Q}{1+Q^2} \quad \cos(\phi) = \frac{1-Q^2}{1+Q^2}$$

Thus, we can remark that Eq. 10 depends only on $\phi$ and is a quadratic equation of Q

$$(\sqrt{2}S_1 Q - \sqrt{2}QC_3 - \sqrt{2}QC_1 S_3 - 2Q + 2 - \sqrt{2}C_1 S_3 - \sqrt{2}C_3 - \sqrt{2}S_1)$$
$$(Q-1) = 0$$

One solution is $Q = 1$, i.e. $\phi = \pi/2 + 2k\pi$ that does not depend on the actuated joints. Figure 12 depicts the four direct kinematic solution for $\theta_1 = 0.1$, $\theta_2 = 0.2$, $\theta_3 = \pi/4$. Solutions (a) and (b) are found when $Q = 1$ and can be easily isolated. From solutions (c) and (d), only the second one is suitable that can be isolated by the dot product of $\mathbf{r}_2$ by $\mathbf{p}_2$.

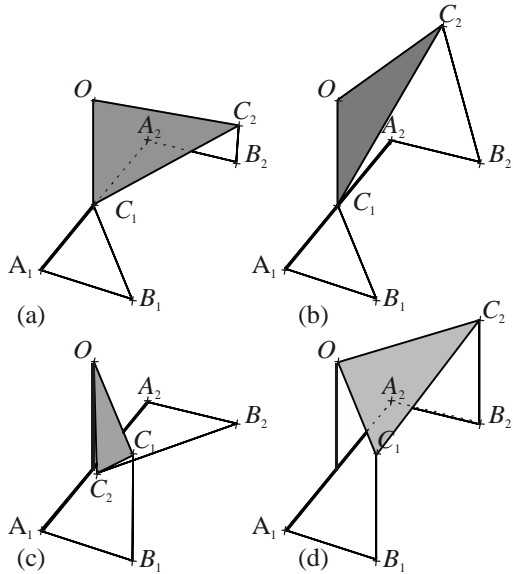

Fig. 12: The four direct kinematic solutions for
$\theta_1 = 0.1$, $\theta_2 = 0.2$, $\theta_3 = \pi/4$

To solve the inverse kinematic, we use two substitutions, $R = \tan(\theta_1/2)$ and $S = \tan(\theta_2/2)$ that permit us to have two quadratic and independent equations as function of $R$ and $S$ respectively. Figure 13 shows the four inverse kinematic solutions for $\theta_3 = \pi/4$, $\phi = \pi/12$, $\psi = \pi/12$ that we can isolate by calculating $\mathbf{l}_1.\mathbf{r}_1$ and $\mathbf{l}_2.\mathbf{r}_2$ for legs ① and ②, respectively.

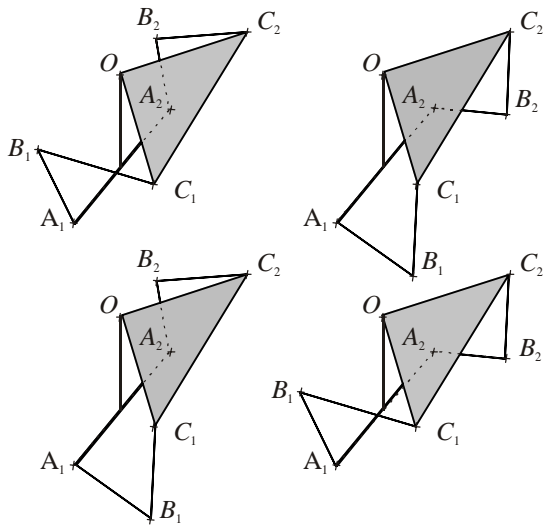

Fig. 13: The four inverse kinematic solutions for
$\theta_3 = \pi/4$, $\phi = \pi/12$, $\psi = \pi/12$

To conclude, we have four solutions for the direct kinematic and four solutions for the inverse kinematic (two for legs ① and ②, respectively).

## IV. CONCLUSIONS

The design of spherical wrists taking into account the constraints on the biomimetic of the eel was made in this paper. A new architecture is investigated and isotropic constraints are applied to produce three suitable solutions. The symmetrical constraints lead us to choose the one where the placement of the actuated joints is optimal because they are located on a median plane where the focal distance is maxima.

### ACKNOWLEDGMENTS

This research was partially supported by the CNRS ("Anguille" Project).